\pdfoutput=1
\documentclass[conference]{IEEEtran}
\IEEEoverridecommandlockouts
\usepackage{cite}
\usepackage{amsmath,amssymb,amsfonts}
\usepackage{graphicx}
\usepackage{wrapfig}
\usepackage{textcomp}
\usepackage{xcolor}
\usepackage{algorithmicx}
\usepackage{algcompatible}
\usepackage[ruled,vlined,linesnumbered]{algorithm2e}
\usepackage{amsthm}
\def\BibTeX{{\rm B\kern-.05em{\sc i\kern-.025em b}\kern-.08em
    T\kern-.1667em\lower.7ex\hbox{E}\kern-.125emX}}
\begin{document}

\hyphenation{}

\newtheorem{remark}{Remark}
\newtheorem{corollary}{Corollary}
\newtheorem{proposition}{Proposition}
\newtheorem{definition}{Definition}
\newtheorem{problem}{Problem}
\newtheorem{theorem}{Theorem}
\newtheorem{lemma}{Lemma}

\title{\textit{e-Uber}: A Crowdsourcing Platform for Electric Vehicle-based  Ride- and Energy-sharing\\
}

\author{\IEEEauthorblockN{Ashutosh Timilsina}
\IEEEauthorblockA{\textit{Department of Computer Science} \\
\textit{University of Kentucky}\\
Lexington, USA \\
ashutosh.timilsina@uky.edu}
\and
\IEEEauthorblockN{Simone Silvestri}
\IEEEauthorblockA{\textit{Department of Computer Science} \\
\textit{University of Kentucky}\\
Lexington, USA \\
simone.silvestri@uky.edu}
}


\maketitle

\begin{abstract}
The sharing-economy-based business model has recently seen success in the transportation and accommodation sectors with companies like Uber and Airbnb. There is growing interest in applying this model to energy systems, with modalities like peer-to-peer (P2P) Energy Trading, Electric Vehicles (EV)-based Vehicle-to-Grid (V2G), Vehicle-to-Home (V2H), Vehicle-to-Vehicle (V2V), and Battery Swapping Technology (BST). In this work, we exploit the increasing diffusion of EVs to realize a crowdsourcing platform called \textit{e-Uber} that jointly enables ride-sharing and energy-sharing through V2G and BST. e-Uber exploits \textit{spatial crowdsourcing}, reinforcement learning, and \textit{reverse auction} theory. Specifically, the platform uses reinforcement learning to understand the drivers' preferences towards different ride-sharing and energy-sharing tasks. Based on these preferences, a personalized list is recommended to each driver through \textit{\underline{C}MAB-based \underline{A}lgorithm for task \underline{R}ecommendation \underline{S}ystem} ($CARS$). Drivers bid on their preferred tasks in their list in a reverse auction fashion. Then e-Uber solves the task assignment optimization problem that minimizes cost and guarantees V2G energy requirement. We prove that this problem is NP-hard and introduce a bipartite matching-inspired heuristic, \textit{\underline{B}ipartite \underline{M}atching-based \underline{W}inner selection} ($BMW$), that has polynomial time complexity. Results from experiments using real data from NYC taxi trips and energy consumption show that e-Uber performs close to the optimum and finds better solutions compared to a state-of-the-art approach.
\end{abstract}


\section{Introduction}
With the recent advent of sharing-economy-based models and their successful application in accommodation-sharing (e.g. Airbnb, Vrbo) and ride-sharing (e.g. Uber, Lyft), researchers have focused on applying this concept to energy systems~\cite{kalathil2017sharing, timilsina2021reinforcement}. Energy-sharing modalities such as peer-to-peer (P2P) energy trading~\cite{timilsina2022prospect, tushar2020peer}, and Electric Vehicle (EV)-based Vehicle-to-Grid (V2G), Vehicle-to-Home (V2H), Vehicle-to-Vehicle (V2V)~\cite{shurrab2021efficient}, as well as Battery Swapping Technology (BST)~\cite{sarker2014optimal} have been proposed as sustainable and flexible approaches to balance the energy supply and demand for both the grid and end-users~\cite{shurrab2021efficient,ai2021crowdsourcing}. Especially, the rapid rise in EV sales in recent years has created new opportunities for mobile and flexible energy storage and management including ride-sharing and energy-sharing services using EVs~\cite{shurrab2021efficient}. However, no studies have been made so far to realize a platform that jointly enables both ride-sharing and energy-sharing. 

Crowdsourcing is an approach for recruiting workers from a ``crowd'' to execute tasks that has been successfully applied to several domains~\cite{restuccia2018incentme,restuccia2018first}. We believe that a crowdsourcing platform has the potential to also be successfully applied to the a combined ridesharing and energy-sharing system, where \textit{tasks} are ride- and energy-sharing requests  that can be performed by EV drivers, called \textit{workers}. Tasks are requested by \textit{task-requesters} which include ride-sharing clients as well as private or public energy customers. Examples of such energy customers include a utility company and a microgrid community looking to achieve demand response by shifting energy demand to V2G services at different locations, specially during the time of peak energy demands~\cite{casella2022hvac,ciavarella2016managing,dolce2018social,timilsina2017technical}.

\begin{figure}[!hp]
    \centering
    \includegraphics[width = 0.85\linewidth]{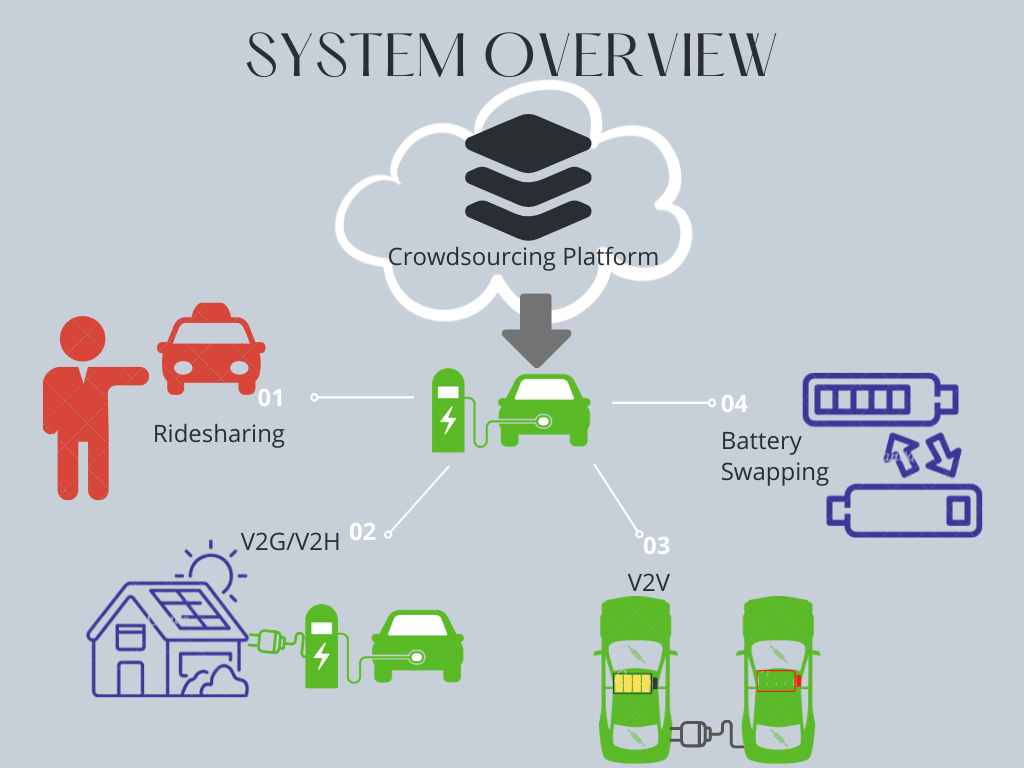}
    \caption{e-Uber crowdsourcing platform overview}
    \label{fig:platform_details}
\end{figure}

In this work, we propose a novel crowdsourcing platform called \textit{e-Uber} that leverages the increasing diffusion of EVs to enable joint ride-sharing and energy-sharing services. A general overview of the platform is depicted in Fig.~\ref{fig:platform_details}. With this platform, drivers equipped with EVs can not only transport passengers through ride-sharing but also sell excess energy stored in their batteries to the grid/houses during periods of high demand through V2G or battery swapping~\cite{khanal2019comparative,timilsina2016design,timilsina2017analysis}.
e-Uber has the potential to increase the earning potential for drivers and also to help balance the energy demand and supply for the grid while simultaneously fulfilling the mobility and energy demands of consumers. 

A few works on crowdsourcing have been proposed to facilitate the integration of energy-sharing services with EVs. Ai et al.~\cite{ai2021crowdsourcing} proposed a V2H-based omni-sharing modality in a microgrid community to crowdsource energy from EVs. 
Similarly, the authors in~\cite{yassine2020cloudlet} propose an autonomous EV (AEV)-based energy crowdsourcing approach, allowing AEVs to participate in energy-sharing tasks for consumers placed in the cloudlet. 
However, these approaches do not consider  the {\em workers' preferences} as well as their {\em limited ability} of selecting tasks  when overwhelmed with choices and problems. There have been a few spatial crowdsourcing work attempting at solving the task assignment problem considering worker preferences \cite{zhao2020preference,restuccia2018first,jin2016enabling,xing2019multi}. However, these approaches focus on general uniform tasks, and do not consider ride-sharing combined with energy-sharing.


To the best of our knowledge, in this paper we propose the first crowdsourcing mechanism that jointly enables ride- and energy-sharing to provide a multifaceted solution to existing problems on efficiency and sustainability of transportation, energy management, and cost-effective demand response using EVs. \textit{e-Uber} works in three decision stages: calculate a personalized task recommendation for each EV worker, collect bids from workers, reverse auction-based winning bids selection. We propose a preference-aware optimal task recommendation system, $POTR$, and a reinforcement learning mechanism to learn worker preferences. The reverse auction process is formalized for bidding and the winning bids are determined through an optimization framework called \textit{Winning Bid Selection} ($WiBS$). A Reinforcement Learning (RL)-based algorithm, called $CARS$ is proposed that solves the problem of task recommendation and updates the worker preferences based on their interaction with the recommendation using Combinatorial Multi-Armed Bandit framework~\cite{timilsina2021reinforcement}. Proving that $WiBS$ problem is NP-hard, we also propose bipartite matching-based heuristic, $BMW$ that finds solution to $WiBS$ in polynomial time. 

The major contributions of the paper are as follows:
\begin{itemize}
    \item We propose a spatial crowdsourcing platform, \textit{e-Uber}, to jointly enable ride-sharing and energy-sharing using EVs;
    
    \item We develop an optimization framework, called \textit{POTR}, based on reinforcement learning for personalized recommendation of tasks to workers. 
    
    \item We also formalize winning bid selection (\textit{WiBS}) problem, and prove that it is NP-Hard;

    \item We propose an RL algorithm, called $CARS$, that incorporates reinforcement learning for task recommendation to workers and update the preferences according to their interaction to the recommendation;

    \item Given the complexity of the $WiBS$ problem, we propose a Bipartite Matching-based Winner Selection algorithm, $BMW$ and determine its polynomial time complexity; 
    
    \item Through extensive experiments using real data, we show that \textit{e-Uber} can indeed lead to successful joint crowdsourcing of energy and ride-sharing services that is able to complete more than 850 tasks compared to state-of-the-art approach in a span of 24 hours; 
\end{itemize}

\section{Related works}
Crowdsourcing services has received increasing attention in recent years because of their flexibility and convenience in facilitating the completion of tasks by a set of  workers \cite{restuccia2018first}. There exists a plethora of research works that focus on different aspects of crowdsourcing from optimal task allocation~\cite{shurrab2021efficient} to preference-aware decision-making~\cite{zhao2020preference} to privacy-preserving~\cite{sodagari2022trends,jin2016enabling}. Some other focus on designing an effective and informed incentive mechanism that motivates workers for their sustained engagement in the system~\cite{jin2016enabling}.
Reverse auction mechanism has been widely utilized for designing incentive mechanism including bidding and winner selection in crowdsourcing works~\cite{xing2019multi,xiao2019sra,liu2019truthful,hong2019crowdsourcing}. In~\cite{xiao2019sra}, a secure reverse auction protocol is devised for task assignment for spatial crowdsourcing along with an approximation algorithm. Similarly,~\cite{liu2019truthful} proposes a truthful reverse auction mechanism for location-aware crowdsensing while authors in~\cite{xing2019multi} focus on generalized second-price auction for stable task assignment. The work in~\cite{hong2019crowdsourcing} also uses a truthful reverse auction mechanism to devise incentives for workers in urban parcel delivery.

In context of electric vehicles (EV), the work in~\cite{shurrab2021efficient} employs crowdsourcing for solving charging problems of EVs. A V2V energy-sharing framework has been proposed that crowdsources the charging request from EV owners and allocates the energy considering energy trading prices, EV parameters and privacy. Some other crowdsourcing literature focus on different problems like route optimization of EVs~\cite{wang2013context} and parcel delivery using EVs~\cite{he2021model}.
Closer to our problem setting, some literature have explored the use of crowdsourcing for integrating energy-sharing services with EVs. For instance, authors in~\cite{ai2021crowdsourcing} proposed a V2H-based omni-sharing modality system in a microgrid community, where energy is crowdsourced from EVs to reduce the overall cost of the community and decrease the need for energy storage. Another study~\cite{yassine2020cloudlet} suggested an autonomous EV-based energy crowdsourcing approach, which enables EVs to participate in energy-sharing tasks for cloud-based energy consumers. However, this approach is challenging to implement and doesn't consider workers' preferences or the impact of sub-optimal decision-making. 

In fact, most of these crowdsourcing works ignore the user behavioral modeling in task assignment. The spatial crowdsourcing work in~\cite{zhao2020preference} tried to solve the task assignment problem by considering worker preferences, but this solution is better suited for group tasks and doesn't account for other behavioral aspects of user behavioral modeling like \textit{bounded rationality}~\cite{kahneman2003maps} and irrational decision-making that drastically affects the system performances. Additionally, the existing works neglect the task recommendation problem and other realistic budget constraints, such as the energy budgets required by the utility or microgrid for any time period. Furthermore, these works are limited to homogeneous tasks like energy-sharing or delivery services only, which can result in significant idle hours for EVs during off-peak periods as such tasks have similar pattern.

In conclusion, while existing literature in crowdsourcing mechanisms have contributed to task assignment, incentive design, privacy and energy-sharing services, there is still room for improvement in terms of behavioral aspect like preference-aware task recommendation and online learning of these preferences; task assignment with overall cost minimization and energy budgets; and heterogeneity in crowdsourcing tasks. Our proposed work focuses on addressing these limitations and developing more comprehensive, effective, and realistic solution to joint enabling of ride-and energy-sharing services in a crowdsourcing setting using reverse auction, reinforcement learning and efficient matching algorithms.

\section{System Model}

\begin{table}[ht]
\centering
\caption{List of Notations}
\label{tab:my-table}
\resizebox{0.99\linewidth}{!}{%
\begin{tabular}{|l|l|}
\hline
$\mathcal{S}_t$  & List of all tasks at timeslot $t$ \\ \hline
$s_j =\langle z_j,c_j, d_j\rangle$ & $j^{th}$ task represented by type of task ($z_j$), start position ($c_j$), \\ & destination ($d_j$) \\ \hline
$\mathcal{W}_t$  & List of all workers available at timeslot $t$ \\ \hline
$w_i=\langle c_i,e_i,r_i,r_i^{min}\rangle$ & $i^{th}$ worker represented by current location ($c_i$), the energy \\ & per unit range ($e_i$), remaining range of the EV ($r_i$) and \\ & minimum range threshold ($r_i^{min}$) \\ \hline
$\mathcal{B}_t$ & List of all the bids received at timeslot $t$ \\ \hline
$b_{ij}$ & Bid submitted by worker $i$ for task $j$ \\ \hline
$\alpha_{iz_j}$ & Acceptance probability of worker $i$ to the $j^{th}$ task type $z_j$ \\ \hline
$K$ & Maximum number of tasks to be recommended \\ \hline
$\lambda$ & Proximity distance \\ \hline
$\mathcal{E}_t$ & Amount of energy that must be satisfied through V2G/V2H \\ \hline
$\mathbf{q}^*$ &  Optimal solution to $f(.)$/ Winning bids \\ \hline
\end{tabular}%
}
\end{table}

We assume time to be divided in time slots. At each time slot $t$, the set of tasks is referred to as $\mathcal{S}_t$, which are crowdsourced to the workers. We refer to $\mathcal{W}_t$ as the set of workers at time $t$. Each task in $\mathcal{S}_t$ is denoted by a tuple $s_j \stackrel{\text{def}}{=} \langle z_j, c_j, d_j\rangle$ where $z_j$ is the type of task ($0-$rideshare, $1-$battery swapping, and $2-$V2G), $c_{s_j}$ is the start position and $d_j$ is the destination 
of task. For energy-sharing tasks, although spatial in nature, start position $c_{s_j}$ is same as destination $d_j$. We assume the utility company submits energy tasks as a result of an {\em energy requirement} $\mathcal{E}$. This is a typical assumption for demand response solutions \cite{casella2022hvac,ciavarella2016managing,dolce2018social}. As a result, the total amount of energy provided by workers through   V2G must be at least $\mathcal{E}$. 
Each worker in $\mathcal{W}_t$ is denoted by a tuple $w_i \stackrel{\text{def}}{=} \langle c_{w_i}, e_i, r_i, r_i^{min} \rangle$, where $c_i$ is the current position of the EV worker $w_i$ which can be different to spatial task location $c_{s_j}$, $e_i$ is the energy per unit range value in ($kWh/km$) that gives information about how much energy the EV consumes to drive a unit distance,  $r_i$ is the available range of electrical vehicle in $km$ given by the remaining energy level in their batteries, and $r_i^{min}$ is the minimum energy not to be exceeded after completing the task to ensure sufficient energy for traveling to a charging location. 
The energy required to perform task $s_j$ by  worker $w_i$ is denoted by $l_{ij}$. 
e-Uber provides that a list of tasks, called {\em recommendation list}, is sent out to each worker. Workers  then submit bids to these tasks. The bid $b_{ij} \in \mathcal{B}$ represents the cost asked by worker $w_i$ to perform task $s_j$, where $\mathcal{B}$  is the set of all the bids submitted by workers. 


\begin{figure}[!htp]
    \centering
    \includegraphics[width=0.85\linewidth]{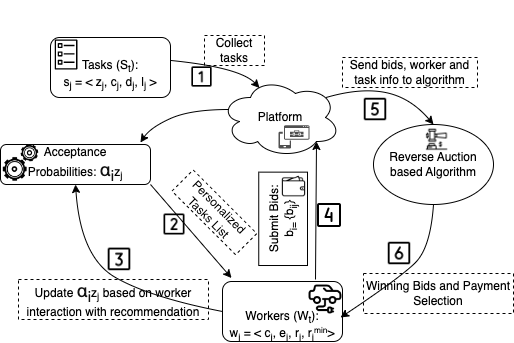}
    \caption{Working mechanism of e-Uber}
    \label{fig:my_label}
\end{figure}

Previous works in crowdsourcing and energy-sharing using EVs has generally assumed that workers would have complete access to the list of available tasks and would pick the best task for them or, conversely, the crowsourcing platform would assign tasks to workers regardless of their preference. These assumptions are both undesirable. On the one hand  workers have limited time and ability to go over potentially a very long list of tasks \cite{timilsina2021reinforcement}, and on the other hand workers may have different preferences on the tasks to complete. In this work,  we recommend a limited list of relevant tasks to each worker based on their preferences.
We model the preferences as follows. We denote by $\alpha_{iz_j}\in [0,1]$ the probability that worker $w_i$ bids for a task of type $z_j$. These are called \textit{bidding probabilities}. We assume that these probabilities are unknown and thus need to be learned over time by observing the workers' behavior.

\section{\textit{e-Uber:} Problem Formulation}\label{sec:problem}
Fig.~\ref{fig:my_label} summarizes the steps involved in the  e-Uber platform. e-Uber collects a list of tasks $\mathcal{S}_t$ at time $t$ as requested by task-requesters which need to be crowdsourced to the EV-based workers in $\mathcal{W}_t$ (step $1$). The platform sends a personalized list of  tasks to the workers based on their preferences (step $2$) to which they respond by submitting bids to the platform for the tasks (step $3$). Based on the received bids $\mathcal{B}_t$ (step $4$), the platform uses reverse auction based algorithm to determine the winning bids $\mathbf{q}^*$ along with final payment $\mathbf{P}$ for winners (step $5$). Finally, the worker preferences are updated based on their feedback for the next time step (step $6$). Given the nature of the considered tasks,  worker-task assignment is performed one-to-one.

As described above, the system involves solving two different problems. One is to recommend a set of tasks which maximizes the likelihood of generating the maximum number of bids, and thus improving the overall system performance. Another problem is to select the winning bids for task assignment and determine the final payment to crowdsource the tasks to the workers. These two problems are discussed below.

\subsection{Preference-aware Optimal Task Recommendation Problem}
Our objective is to recommend a limited subset of tasks to each workers which maximizes the likelihood of bidding for these tasks, while avoiding to overwhelm workers with a list above their cognitive capabilities. We formalize this through the Preference-aware Optimal Task Recommendation (POTR) problem as follows. In short, the problem aims at maximizing the overall task bidding probabilities (hereafter referred interchangeably as preferences) while limiting the size of the recommended list to $K$ as well as ensuring that each task is recommended to at least $\psi$  workers.

\begin{subequations}\label{obj_func_rs}
\begin{align}
	{\text{maximize}}& && \sum_{w_i \in \mathcal{W}}\sum_{s_j \in \mathcal{S}} \alpha_{iz_j} x_{ij} \tag{\ref{obj_func_rs}}\\
	\mbox{s.t.}& && \sum_{s_j\in\mathcal{S}} x_{ij} \leq K, &&&& \forall w_i\label{const1_rs}\\
        &&& \sum_{w_i\in\mathcal{W}} x_{ij} \geq \psi, &&&& \forall s_j\label{const1a_rs}\\
        &&& \sum_{s_j\in\mathcal{S}} g(z_j) x_{ij} \geq \frac{|V2G|}{|\mathcal{S}|}K, &&&& \forall w_i\label{const1b_rs}\\
 	&&& l_{ij}x_{ij} \leq (r_i-r_i^{min}) e_i, &&&&\forall w_i,s_j\label{const2_rs}\\
 	&&& x_{ij} = 0,\ \text{if }|c_{s_j} - c_{w_i}| > \lambda, &&&&\forall w_i,s_j\label{const3_rs}\\
 	&&& x_{ij} \in \{0, 1\}, &&&&\forall w_i,s_j\label{const7_rs}
\end{align}
\end{subequations}

\begin{equation}
g(z_j) =
    \begin{cases}
    1,& \text{if } z_j =2\\
    0,& \text{otherwise}
\end{cases}
\end{equation}

The objective function  in Eq.~\eqref{obj_func_rs} maximizes the sum of individual bidding probabilities for each worker's recommended tasks. The binary decision variable $x_{ij} \in \{0,1\}$ is set to 1 if the task $s_j$ is included in the list of worker $w_i$. Constraint~\eqref{const1_rs} limits the length of each recommendation list to be less than $K$. In constraint~\eqref{const1a_rs},  we ensure that each task is recommended to at least $\psi = \Big\lfloor\frac{|\mathcal{W}|K}{|\mathcal{S}|}\Big\rfloor$ workers. 
Also, we ascertain that a minimum of $\frac{|V2G| \times K}{|\mathcal{S}|}$ V2G tasks are also recommended to each workers in constraint~\eqref{const1b_rs}.
Constraint~\eqref{const2_rs} requires the recommended tasks to consume no more than certain energy for each EV, ensuring that EV has sufficient energy after performing tasks to drive to charging location, if required. Finally, constraint~\eqref{const3_rs} ensures that only the tasks within $\lambda$ distance from workers are recommended. 

It is to be noted that the information on bidding probabilities is difficult to obtain \textit{a priori} as it is specific for each worker and include elements of complex human psychology. Therefore, we assume that the preferences are initially unknown and are  learned by observing the workers' behavior with respect to the assigned tasks.  Recently, reinforcement learning mechanisms have been used extensively to learn the optimal policies in the run-time that gradually converge to take optimal actions based on feedback from the environment. In section~\ref{sec:RL_eUber}, we present a \textit{Combinatorial Multi-Armed Bandit (CMAB)}-based approach~\cite{timilsina2021reinforcement} that learns the preferences of workers over time while simultaneously recommending the optimal personalized list of tasks to them.

\subsection{Winning Bid Selection and Final Payment Problem}
After sending the personalized list of  tasks to each worker, \textit{e-Uber} collects the bids. Given the collected bids, e-Uber selects winning bids, i.e., the workers performing the tasks, by solving the Winning Bid Selection ($WiBS$) problem. This problem determines the best bids which minimize the total cost from perspective of task requesters.  $WiBS$  can then be formulated a costrained assignment problem as follows:
\begin{subequations}\label{obj_func_EV}
\begin{align}
	{\text{minimize}}& && 
	\sum_{w_i\in\mathcal{W}}\sum_{s_j\in\mathcal{S}}b_{ij}q_{ij} \tag{\ref{obj_func_EV}}\\
	\mbox{s.t.}& && \sum_{s_j\in\mathcal{S}}  q_{ij} \leq 1, &&&& \forall w_i\label{const1_EV}\\
 &&& \sum_{w_i\in\mathcal{W}} q_{ij} = 1, &&&& \forall s_j ,z_j < 2\label{const2_EV}\\
  &&& \sum_{w_i\in\mathcal{W}} q_{ij} \leq 1, &&&& \forall s_j ,z_j = 2\label{const3_EV}\\
	&&& \sum_{w_i\in\mathcal{W}}\sum_{s_j\in\mathcal{S}}g(z_j)l_{ij}q_{ij} \geq \mathcal{E}, 
	\label{const4_EV}\\
 	&&& q_{ij} \in \{0, 1\}, &&&&\forall w_i,s_j\label{const7_EV}
\end{align}
\end{subequations}




The objective function in Eq. \eqref{obj_func_EV} minimizes the total cost of performing tasks from the collected bids. $q_{ij}$ is the binary decision variable as defined in constraint \eqref{const7_EV} that indicates whether a bid $b_{ij}$ wins the auction and therefore the task $s_j$ is assigned to worker $w_i$. Constraint \eqref{const1_EV} ensures that a worker is assigned at most one task, while  
\eqref{const2_EV} allows a ride-sharing and battery swapping tasks ($z_j <2$)  to be assigned to only one worker. Similarly, constraint \eqref{const3_EV} ensures that a V2G task is assigned to at most one worker.  Finally, constraint \eqref{const4_EV}, ensures that at least $\mathcal{E}$ amount of energy will be supplied through V2G services. Note that the function $g(z_j) =  1$ if $z_j = 2$ (V2G task) and zero otherwise.  

Following the selection of winning bids by solving the $WiBS$ problem in Eq. \eqref{obj_func_EV}, the final payment for each winning worker $w_k$ assigned with task $s_j$ is the second-to-the-selected bid received for that task. 
Since with the second price payment rule, the dominant strategy for all bidders is to bid truthful~\cite{lazar1998progressive}, it ensures rational workers will provide truthful bids. 

\begin{theorem}\label{th:np-hard}
$WiBS$ problem defined in Eq. \eqref{obj_func_EV} is NP-hard.
    \begin{proof}
We provide a reduction from NP-Hard 0-1 min Knapsack (0-1 min-KP) problem~\cite{csirik1991heuristics}. In this problem, a set $n$ items is provided, each item $a_i$ has a value $l_i$ and weight $b_i$. The goal is to select the subset of items that incurs minimum weight and has a value of at least $\mathcal{E}$.

Given a generic instance of min-KP, we construct an instance of our problem as follows. We only consider V2G tasks ($z_j = 2$). For each item $a_i$ of min-KP we create a pair task-worker $(s_{a_i},w_{a_i})$. We assume that worker $w_{a_i}$ only submits one bid, and they bid for $s_{a_i}$ for an amount $b_i$ (the weight of $a_i$ in min-KP). Additionally, the energy required by $w_{a_i}$ to perform $s_{a_i}$ is $l_i$ (the value of $a_i$ in min-KP). Finally, we set the energy requirement for V2G to $\mathcal{E}$.

Under these assumptions, the decision variable $q_{ij}$ of our original problem can be reduced to $q_i$, since only one workers bid for one task and a task receives a bid only from one worker. Additionally, constraints \eqref{const1_EV} and \eqref{const3_EV} are trivially verified, since there is only  task-worker pair, while constraint \eqref{const2_EV} does not apply since we only have V2G tasks. 

Solving our reduced problem instance finds the set task-worker pairs that minimize the sum of bids and meets the energy requirement  $\mathcal{E}$. This corresponds (i.e., it can be translated in polynomial time) to the optimal solution of min-KP, i.e., the set of items with minimum weight that provide a value at least  $\mathcal{E}$. As a result, our problem is at least as difficult as min-KP, and thus it is NP-Hard.

    \end{proof}
\end{theorem}

\section{e-Uber Solution Approaches}\label{sec:RL_eUber}
\subsection{CMAB-based Task Recommendation System}
In order to solve the optimization problem in Eq. \eqref{obj_func_rs}, it is necessary to have beforehand knowledge on the workers preferences. These are generally {\em not known} a priori in realistic settings. Therefore, it becomes necessary to learn these preferences during run-time, 
while simultaneously optimizing the task assignment. To this purpose, 
we propose a reinforcement learning approach inspired by the Combinatorial Multi-Armed Bandit (CMAB) framework ~\cite{gai2012combinatorial,timilsina2021reinforcement}. 

Combinatorial Multi-Armed Bandit is a classic reinforcement learning problem that consists of setup where agents can choose a combination of different choices (i.e. certain decision-making \textit{actions}) and observe a combination of linear \textit{rewards} at each timestep. The long term objective for the problem is to find a strategy that maximizes such reward by selecting optimal actions. This strategy, better defined as \textit{policy}, needs to be learned based on how the agents choose to interact with the system. The learning is carried out through \textit{exploration vs. exploration} trade-off. Since, at the beginning, the knowledge about how an agent chooses to engage with the system is not known, the system learns by allowing agent to choose from diverse options and therefore learning the user interaction accordingly, referred to as \textit{exploration}. As the time passes, the system starts gathering information about agent's behavior and therefore use that knowledge instead of sending out diverse range of choices, called \textit{exploitation}. By balancing this exploration and exploitation mechanism over the course of time, the system eventually gathers sufficient information on agent's behavior and learns optimal strategy for them. In our problem setting, the workers are the agents who needs to be sent out an optimal set of tasks so as to accumulate good quality bids from them. Specifically, the objective is to find the best possible task recommendations (actions) to be sent to each workers (agent) that will result in higher cumulative preferences for workers (reward).


Therefore in this section, based on this CMAB framework, we design an algorithm called \textit{\underline{C}MAB-based \underline{A}lgorithm for task \underline{R}ecommendation \underline{S}ystem (CARS)}. The pseudo code of \textit{CARS} is shown in Alg. \eqref{alg_1}. \textit{CARS} recommends the personalized tasks to each workers based on current estimation of worker preferences towards each task type. Note that the worker preference is defined as the bidding probability in section~\ref{sec:problem} that a worker will submit a bid for any task based on its type. The algorithm then updates and learns these biding probabilities based on the worker's engagement on the recommendation through bids. If the worker submits a bid, it is considered to be a preferred recommendation and opposite, if the worker chooses to ignore by not the submitting bid. Based on this information, the preference of workers towards each task type is updated. 

Therefore, with $\mathcal{F}$ as the overall solution space that consists of all feasible action matrices, the action matrix $\mathbf{A}(t) \in \mathcal{F}$ corresponds to the optimal set of recommendation lists for the timestep $t$. It consists of action values $x_{ij} \in \{0, 1\}$, which is same as the decision variable in POTR problem. Recall that it represents whether the task $s_j$ is in personalized recommendation list of worker $w_i$ for timestep $t$. Given this action matrix, the preference of worker $w_i$ towards each task type $z_j$ is modeled as a random variable $\Bar{\alpha}_{iz_j}$ whose mean value is $\alpha_{iz_j}$ and is initially unknown. The current knowledge until timestep $t$ for these random variables $\Bar{\alpha}_{iz_j}$ is denoted by the estimated expected $\widehat{\alpha}_{iz_j}$. The reward for the platform for selecting the action matrix $\mathbf{A}(t)$ at timestep $t$, is defined as the sum of the preferences to each  workers:
\begin{equation}\label{eq:reward}
\mathbf{R}_{\mathbf{A}(t)}(t) = \sum_{w_i,s_j} a_{ij}(t)\Bar{\alpha}_{ij}(t)
\end{equation}

Since the distribution of $\Bar{\alpha}_{iz_j}$ is unknown, the goal of this CMAB-based approach is to learn the policy, that minimizes the overall \textit{regret} up to time $t$. This regret is defined as the difference between expected reward with perfect knowledge of preferences and that obtained by the policy over time:

\begin{equation}\label{eq:regret}
\mathcal{R}(t) = t \mathbf{R}^*_{\mathbf{A}(t)}(t) - \mathbb{E}\Big[\sum_{t' = 1}^t \mathbf{R}_{\mathbf{A}(t')}(t')],
\end{equation}

where $\mathbf{R}^*_{\mathbf{A}(t)}(t)$ is the optimal reward obtained with perfect knowledge of the preference variables. Even though minimizing the regret is a difficult problem, $CARS$ ensures that the regret is bounded, meaning the non-optimal actions will be picked only a limited number of times and eventually the learned policy will converge towards optimal. We present a modified objective function from UCB1 algorithm to select the action matrix as follows.

\begin{equation}\label{eq:rl_obj}
    \mathbf{A}(t) = \arg\max\limits_{\mathbf{A} \in \mathcal{F}} \sum\limits_{w_i \in \mathcal{W}} \sum\limits_{s_j \in \mathcal{S}} a_{ij} \left({\widehat{\alpha}_{iz_j} + \sqrt{\frac{(Q+1)\ln t}{ m_{iz_j}}}}\right)
\end{equation}
where $Q = |\mathcal{W}|\times |z_j|$ is the total number of variables and $m_{iz_j}$ is the number of observations so far for the variable $\Bar{\alpha}_{iz_j}$.

At each timestep $t$, we solve the $POTR$ problem with CMAB-based objective function in Eq. \eqref{eq:rl_obj} instead of Eq.~\eqref{obj_func_rs} and same constraints \eqref{const1_rs}-\eqref{const7_rs}. By solving this modified problem, the sets of optimal actions (or recommendation lists) for each workers are selected based on current estimate of preferences until timestep $(t-1)$. For this purpose, we keep track of the 
$\widehat{\alpha}_{iz_j}$, along with $m_{iz_j}$. These two information are then used to update the current estimation of the variable $\Bar{\alpha}_{iz_j}$ at time $t$ 
based on the worker's engagement with the recommendation i.e. whether the worker chooses to submit the bid or not. Needs to be noted that, if the worker chooses to submit the bid, they must complete the task if assigned.

\begin{align}
\footnotesize
 \widehat{\alpha}_{iz_j}(t) &=
 \begin{cases}
  \frac{\widehat{\alpha}_{iz_j}(t-1)m_{iz_j}(t-1)+\alpha_{iz_j}(t)}{m_{iz_j}(t-1)+1} \quad   \text{if $0 < b_{ij} < \infty$,} \\
 \widehat{\alpha}_{iz_j}(t-1)      \qquad \qquad \qquad \text{otherwise.}
 \end{cases}\label{eq:theta}\\
 m_{iz_j}(t) &= m_{iz_j}(t-1)+1 
 \label{eq:mij}
\end{align}

We present the \textit{CARS} algorithm in Alg.~\ref{alg_1}. CARS begins by collecting information on workers and task in lines $1-2$. It then sends out personalized recommendation to each worker by solving the optimization problem with Eq.~\eqref{eq:rl_obj} as objective function and constraints \eqref{const1_rs}-\eqref{const7_rs}(lines $3-4$). Then, it collects the bids for recommended tasks from workers (line $5$). Finally, the current knowledge on worker's bidding probabilities are updated according to the Eqs.~\eqref{eq:theta} and~\eqref{eq:mij} based on how the workers respond to recommendations (lines $5-6$). For the update process, the recommendations that receive a bid from workers are taken as positive reinforcement and the recommendations that do not receive bids as negative reinforcement. In the following, we prove that the Alg.~\ref{alg_1} has a bounded regret and thus the algorithm eventually converges to optimal policy in finite time-steps.

\begin{algorithm}
\SetAlgoLined
\footnotesize

$\forall w_i \in \mathcal{W}_t$, collect the workers info $w_i = <c_i, e_i, r_i, r_i^{min}>$ \; 
$\forall s_j \in \mathcal{S}_t$, collect the tasks $s_j = <z_j,c_j, d_j>$\; \tcc{\bf Solve CMAB-based POTR problem}
Select an action $\mathbf{A}$ s.t.
$\mathbf{A}(t) = \arg\max\limits_{\mathbf{A} \in \mathcal{F}} \sum\limits_{w_i} \sum\limits_{s_j} a_{ij} \left({\widehat{\alpha}_{iz_j} + \sqrt{\frac{(Q+1)\ln t}{ m_{iz_j}}}}\right)$\;
Send list of recommendations $\mathbf{A}(t)$ to the workers\;
Collect bids $\mathcal{B}_t$ from workers based on $\mathbf{A}(t)$\;
Update $[\widehat{\alpha}_{iz_j}]_{|\mathcal{W}|\times |z_j|}$ and $[m_{ij}]_{|\mathcal{W}|\times |z_j|}$ based on the collected bids using Eqs. \eqref{eq:theta} and \eqref{eq:mij}\;
 \caption{CMAB-based Algorithm for task Recommendation System (CARS)}\label{alg_1}
\end{algorithm}

\begin{theorem}\label{theo_2}
eCARS provides bounded regret given by:
\begin{equation}
 \mathcal{R}(t) \leq \left[\frac{4a_{max}^2Q^3(Q+1)\ln(t)}{(\Delta_{min})^2}+\frac{\pi^2}{3}Q^2+Q\right]\Delta_{max},
\end{equation}
where, $a_{max}$ is defined as $\max\limits_{\mathbf{A}\in\mathcal{F}}{\max\limits_{i,j}{a_{ij}}}$. Besides, $\Delta_{min}=\min\limits_{\mathbf{R}_{\mathbf{A}}<\mathbf{R}^*}\left({\mathbf{R}^*-\mathbf{R}_{\mathbf{A}}}\right)$ and $\Delta_{max}=\max\limits_{\mathbf{R}_{\mathbf{A}}<\mathbf{R}^*}\left({\mathbf{R}^*-\mathbf{R}_{\mathbf{A}}}\right)$ are the minimum and maximum difference to the reward obtained with perfect knowledge of the users' preferences, respectively.
\end{theorem}
\begin{proof}
The proof is obtained following Theorem 2 of \cite{gai2012combinatorial}.
\end{proof}

However, as shown in Theorem \eqref{th:np-hard}, finding optimal solution for winner determination problem ($WiBS$ problem Eqs.~\eqref{alg_2}-\eqref{const7_EV}) is NP-Hard problem. Therefore, we devise a bipartite matching-based heuristic for winning bid determination with polynomial time complexity 
for worker-task assignment.


\subsection{Winning Bid Selection using Weighted Bipartite Matching}

The $WiBS$ problem formulation in Eq. \eqref{obj_func_EV} is an extension of one-to-one weighted matching. However, this matching has to select minimum weighted edges for task allocation with energy budget constraints for V2G tasks. Therefore, we hereby develop a heuristic inspired by bipartite minimum weighted matching which can be solved in polynomial time using Karp's algorithm~\cite{karp1980algorithm}. To satisfy the energy budget constraint, we employ iterative matching that removes the highest weighted edges from the previous matching until the budget is met. Simply put, the algorithm runs the minimum weighted matching and if it does not satisfy the budget constraints, removes first $z$ highest weighted edges connected to non-V2G tasks from the previous matching and then runs another round of matching until the feasible solution is found.

\begin{algorithm}[htbp]
\SetAlgoLined
\footnotesize
\SetKwInOut{Input}{Input}
\SetKwInOut{Output}{Output}
\Input{Sets of Workers ($\mathcal{W}$) and  Spatial Tasks ($\mathcal{S}$), Bids ($\mathcal{B}$)}
\Output{Winning bids with final pay ($\mathbf{P}$)}

\tcc{\bf Initialization}
  $\Phi_{out} = \{ \mathcal{W} \cup \mathcal{S}, E_{\Phi} = \emptyset \};\Phi_{temp} = \emptyset; P = \emptyset $ \;
  
  \tcc{\bf Generate bipartite graph $G$}
  $\forall s_j \in \mathcal{S}$, \lIf{$g(z_j) = 1$}{$V \leftarrow \{s_j\}$ \textbf{else} $R\leftarrow \{s_j\}$}
  $\forall w_i \in \mathcal{W}$, collect their respective bids $\mathcal{B}_i$ \; 
  
  
  Build Bipartite Graph $G = \{ \mathcal{W} \cup \mathcal{S}, E_G = \emptyset \}$ \;
  
  \For{each $w_i \in \mathcal{W}, s_j \in \mathcal{S}$}{
  \lIf{$b_{ij} > 0$}{Add edge $(w_i,s_j)$ to $E_G$ with weight, $b_{ij}$}
   
  }   
  \tcc{\bf Run minimum weighted bpt matching until termination}
\While{$\sum\limits_{(w_i,s_j)\in E_{out}}g(z_j)l_{ij} < \mathcal{E}$ or $\Phi_{temp}\neq \Phi_{out}$}{
  $E_{out}\leftarrow$Perform Minimum Weighted Bipartite Matching on $G$\; 
  Output graph $\Phi_{out}= \{ \mathcal{W} \cup \mathcal{S}, E_{out} \}$, where $E_{out} \subseteq E_G$ \;

\tcc{Remove edges if V2G energy budget is not met, and run MWBM on reduced $G$ again}
  \If{$\sum\limits_{(w_i,s_j)\in E_{out}}g(z_j)l_{ij} < \mathcal{E}$}{
  $Z\leftarrow$Select the first $z$ highest weight edges $\in \Phi_{out}$ and $R$ s.t. $\Big(\sum\limits_{(w_i,s_j)\in E_{out}}g(z_j)l_{ij} + \sum\limits_{(w_i,s_j) \in Z}l_{ij} \Big) \geq \mathcal{E}$\;
  \lIf{$Z\neq\emptyset$}
  {Remove all edges $\in Z$ from $G$ and $\Phi_{out}$ \textbf{else} $\Phi_{temp}=\Phi_{out}$} 
  }}
  $\mathbf{q}^* = E_{out}$\;
  \tcc{\bf Final Payment and Task Assignment}
  $\forall w_k \in \mathcal{W}, P_k \leftarrow$ Second to the selected bid $b_{kj}$\;
  Assign the tasks to winning workers along with final price $\mathbf{P}$\;
\caption{\underline{B}ipartite \underline{M}atching-based \underline{W}inner selection (BMW)
}\label{alg_2}
\end{algorithm}


This algorithm called \textit{Bipartite Matching-based Winner selection (BMW)} is presented in Alg.~\ref{alg_2}. $BMW$ takes set of available workers $\mathcal{W}$, tasks $\mathcal{S}$, and the set of bids $\mathcal{B}$ as input and finds the winning bids with final pay $P$ as the output. In line $1$, the algorithm initializes the output graph $\Phi_{out}$, a temporary graph $\Phi_{temp}$ for iterative matching purpose, and $P$. Then it creates a separate sets for V2G and non-V2G tasks as sets $V$ and $R$ in line $2$ and collects the bids from all workers (line $3$). With the information on bids, $BMW$ generates a bipartite graph $G$ between bipartite sets of workers $\mathcal{W}$ and tasks $\mathcal{S}$, and adds edges between those nodes that have non-zero bids i.e. worker $w_i$ with non-zero bid $b_{ij}$ is connected with task $s_j$ (lines $4-7$). Now, it runs a bipartite matching iteratively with while loop in lines $8-15$. Initially, both of the conditions for while loop are true and therefore the algorithm runs first round of Minimum Weighted Bipartite Matching on graph $G$ (line $9$). It then assigns the matched graph to the output graph $\Phi_{out}$ (line $10$) and checks if the energy budget for V2G tasks is satisfied (line $11$). If it is met in the first round, it breaks out of the while loop and determines final payment and task assignment. If it is not met, BMW removes the first $z$ highest weighted edges in $\Phi_{out}$ from $G$ that just meet the remaining of energy budget not met (line $12-13$). Then, since both of the conditions are still true, the algorithm runs another round of matching on reduced graph $G$. 
Eventually the final matching in output graph $\Phi_{out}$ is used as winning task assignments with final payment as per the bid (line $16-18$).

\begin{theorem}\label{Theo:complexity}
The time complexity of the $BMW$ algorithm is $O(|\mathcal{W}|.|\mathcal{S}|^2.log(|\mathcal{S}|))$. 
\end{theorem}
\begin{proof}
The complexity is dominated by the  $while$ loop (lines $10-17$), executed at max $|\mathcal{S}|$ times. 
It involves running minimum weighted full matching as presented in \cite{karp1980algorithm}, which has run time of $O(|\mathcal{W}|.|\mathcal{S}|.log(|\mathcal{S}|))$. Therefore, the overall complexity of the BMW is $O(|\mathcal{W}|.|\mathcal{S}|^2. log(|\mathcal{S}|))$.
\end{proof}

\section{Experiment}
In this section, we present the experimental details for the proposed system, comparison approaches and detailed study of performance of the algorithms.

\subsection{Experimental Setup}

Our experimental setup consists of modeling workers, tasks and the simulation platform. In case of workers, we gathered the publicly available data on $54$ different EV models on battery size, range, charging power and charging speed, and formulated an individual profile for each EV in concern. Similarly for ride-sharing tasks, the high volume taxi trip data of New York City (NYC) from the year of 2013 \cite{nycTlcDataset} was used. The V2G tasks were generated from the 15 minutes energy consumption data from 25 NYC residences from PecanStreet \cite{EnergyDataset}. In absence of real dataset on battery swapping tasks, half of the ride-sharing tasks were extracted as the battery swapping tasks, given their similar profile with batteries transported instead of passengers. These tasks are spatial, therefore, we collect the information on locations, distance, and time required to complete the tasks. 

Furthermore, the simulation platform, e-Uber for crowdsourcing is developed using Python and Gurobi, NetworkX, and PyTorch libraries. We consider a reverse auction period resolution of 15 minutes which corresponds to the standard set by grid for energy trading. This means that every $15$ minutes the e-Uber algorithm will gather the tasks, push the personalized list of tasks to workers, collect the bids and assign the tasks to EV workers that minimizes the overall cost for the task requesters. We set the search radius for the tasks $\lambda = 10$ km and the maximum length of recommendation list $K=5$. The energy budget for each $15$ minutes time period was considered to be total of all $25$ V2G tasks available. The user preferences were sampled uniformly from the set $\{ 0.1, 0.4, 0.5, 0.7, 0.9, 1.0\}$. The energy, time and location of the EVs are tracked and updated accordingly so as to simulate their real-world trip behavior. If the battery level of the cars fall below minimum level, they are considered for the charging for the next time-step.

For comparison approach, we use the task-centric winner selection algorithm as presented in~\cite{liu2019truthful} and refer it as $BG$ for baseline greedy. This approach neither considers user-preference in the problem-setting nor it considers the personalized recommendation system. So for comparison purpose, we augment this method with perfect knowledge-based recommendation system that pushes $K$ best tasks as recommendation to each workers. 
Then we implement the algorithm as presented in~\cite{liu2019truthful} that sorts the bids from lowest to highest for each tasks and assigns them one by one. Note that this approach may not guarantee a complete matching between workers and tasks as the tasks that are processed towards the end may not have any workers left to choose from because of limited number of bids and greedy selection approach. We use this $BG$ as our baseline and compare the performance of our algorithms $CARS$ and $BMW$ along with their perfect knowledge variation $PK$ which has the perfect knowledge on the worker preferences and thus do not involve learning, and $OPT$ optimal solution to $WiBS$ problem. The ride-share dataset in concern consists of actual ride-fare for specific car. However, we require bids from each vehicle for recommended tasks and a realistic model for bid generation is quite difficult to obtain. Therefore we trained a Deep Neural Network with existing dataset for determining the ride fare of the given ride-sharing tasks, the details of which is presented in the following.

\subsection{Results}

\begin{figure}[htpb]
\minipage{0.50\linewidth}
    \centering
    \includegraphics[width=.99\linewidth]{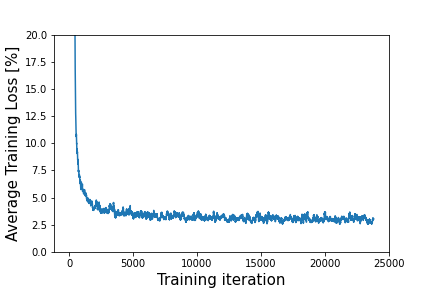}
    \caption{Training Loss \% }
    \label{fig:bids_train}
\endminipage\hfill
\minipage{0.50\linewidth}
    \centering
    \includegraphics[width=.99\linewidth]{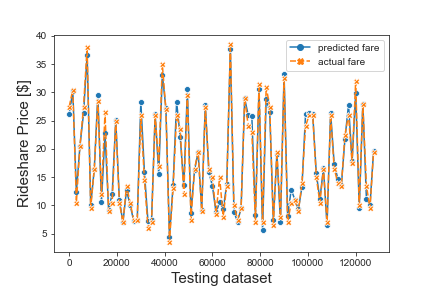}
    \caption{Bid Prediction testing accuracy}
    \label{fig:bids_test}
\endminipage
\end{figure}

\begin{figure*}[!htpb]
  \minipage{0.33\linewidth}%
  \includegraphics[width=.99\linewidth]{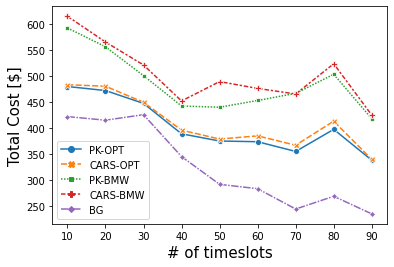}%
  \caption{Snapshot of obj. values \& matches vs. time}\label{fig:obj_match}
  \endminipage\hfill
  \minipage{0.33\linewidth}%
 \includegraphics[width=.99\linewidth]{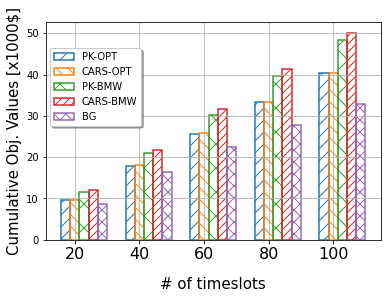}
 \caption{Cumulative obj. values}\label{fig:cum_obj_val}
\endminipage\hfill
  \minipage{0.33\linewidth}%
 \includegraphics[width=.99\linewidth]{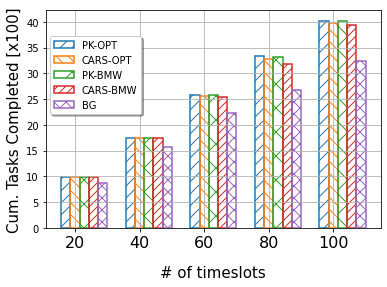}
 \caption{Cumulative tasks}\label{fig:cum_task}
\endminipage\hfill
\end{figure*}

\noindent\textbf{Bid Generation DNN Model}

We used 11 months of taxi data to train and test the DNN model with 80-20 train-test split. The DNN model consisted of 3 hidden layers of sizes ($132,132,64$). We employed ReLU activation function as well as one-hot encoding for the input features, and set the learning rate to 0.0001. The training was carried out for $3$ epochs with $7974$ training batches and batch size of $64$. Consequently, the average training loss curve presented in Fig.~\ref{fig:bids_train}, shows that the loss percentage reduces to $\sim2.5\%$ after $\sim12,000$ trainings. On testing dataset, the bid generation DNN model, generated highly accurate fare prediction with $96.45\%\ R^2-$score. This can also be observed in Fig.~\ref{fig:bids_test} which presents a plot of sample of prediction fares and actual fares to show testing accuracy. 

This DNN model was then deployed in conjunction with the e-Uber to simulate the bidding action by each workers for each recommended tasks in the personalized list. In case of V2G tasks, the energy to be supplied by the EV was converted into its distance equivalent and fed into the DNN model along with other input features to get the bids.

\noindent\textbf{Experimental Observations}

\begin{figure*}[tbhp]
\minipage{0.33\linewidth}
    \centering
    \includegraphics[width=.99\linewidth]{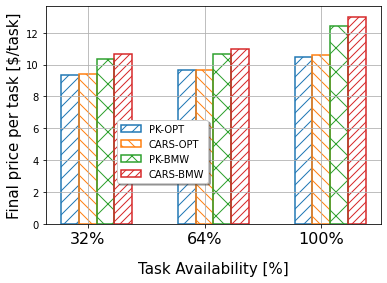}
    \caption{Avg. Price/task vs. Task(\%)}
    \label{fig:fin_price_both}
\endminipage\hfill
\minipage{0.33\linewidth}
    \centering
    \includegraphics[width=.99\linewidth]{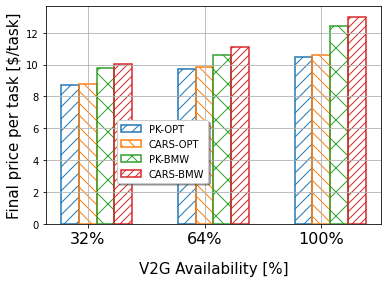}
    \caption{Avg. price/task vs. V2G (\%)}
    \label{fig:fin_price_V2G}
\endminipage\hfill
\minipage{0.33\linewidth}
        \centering
    \includegraphics[height=0.72\linewidth,width= 0.95\linewidth]{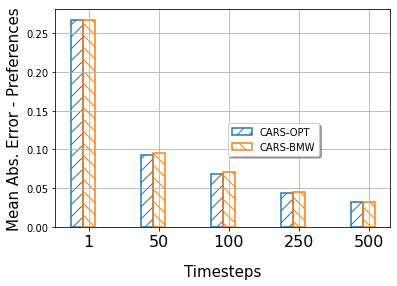}
    \caption{Mean Absolute Error  vs. time}
    \label{fig:sMAPE}
\endminipage\hfill
\end{figure*}

\noindent 1. \textbf{\underline{Performance over time -- Total Cost \& \# of Tasks}}: In the first experimental scenario, we observe the performance of algorithms as a snapshot of objective values over 24 hours (i.e. $24\times4=96$ timeslots). We present the objective values from midnight to next midnight as a lineplot in Fig.~\ref{fig:obj_match} and cumulative bar plots of objective values (Fig.~\ref{fig:cum_obj_val}) and total tasks completed (Fig.~\ref{fig:cum_task}) over a day. Although all the proposed approaches start from the same initial state (except for knowledge on preference), these algorithms may have different successive states since the solution is affected by the matching in previous timeslot, availability of specific workers for next round, and the distance travelled by these workers for previous assignment (or next assignment). Therefore, we employ cumulative objective values and cumulative tasks completed as the metric for a fair comparison of the approaches in Fig.~\ref{fig:cum_obj_val}. This cumulative objective value reflects the overall quality of task assignment made so far based on the total objective values to achieve the requirement while the cumulative tasks completed present the total number of matches made by the respective approach until the end of that timeslot. 
As seen in the lineplot Fig.~\ref{fig:obj_match} and barplot Fig.~\ref{fig:cum_obj_val}, the solution generated by baseline greedy approach $BG$ is the minimum one as it assigns task based on respective cheapest bid available but it doesn't meet the maximum number of matching possible unlike other approaches as shown in Fig.\ref{fig:cum_task}. Therefore, $BG$ mostly violates the V2G requirement, meaning it generates infeasible solutions and hence fails for this problem setting. The $PK-OPT$ produces the best result since it involves solving the $POTR$ and $WiBS$ problem optimally with perfect knowledge of the worker preferences. Following it, is the optimal solution $OPT$ paired with our proposed learning framework for e-Uber, $CARS$, which performs close to optimal in terms of both objective values and number of tasks completed. Although this approach $CARS-OPT$ finds optimal solution, it does not have initial knowledge on preferences. Therefore, it generates sub-optimal recommendation list which then affects the solution to $WiBS$ problem and hence, the overall performance. However, even with online learning framework employed, it produces similar results to the $PK-OPT$. Also we observe similar pattern with $PK-BMW$ and $CARS-BMW$ since they both rely on bipartite matching-based approach to find feasible solution. Since $PK-BMW$ sends the optimal recommendation to workers for collecting bids, it therefore has higher overall performance compared to $CARS-BMW$ which learns the preferences over time. The gaps between best performing $PK-OPT$ and worst performing $CARS-BMW$ however is less than $\$150$ which amounts to a price hike of $\sim\$3/$task in the worst case with an average $50$ tasks for a timeslot as in our case. 
We 
observe the cumulative objective values grow almost linearly for all approaches 
and as expected, the performance observed was better for $PK-OPT$ followed by $CARS-OPT$ and then $PK-BMW$ and finally $CARS-BMW$. However, the gap in cumulative objective value increased for the bipartite heuristic compared to optimal due to its sub-optimal performance. Note that the baseline $BG$ generates less cumulative objective value but it fails to generate maximal matching as seen in Fig.~\ref{fig:cum_task}. The number of tasks completed by the proposed approaches exceed 850 more than the $BG$ in the span $24$ hours.

\noindent 2. \textbf{\underline{Average final price per task and scaling}}: In this experiment, we track the average final price per task while scaling the available tasks from $32\%$ to $64\%$ and then at $100\%$. For scaling the tasks, we increase the number of each type of tasks proportionally. The result is plotted in Fig. \ref{fig:fin_price_both}. 
As the system scales, the average final price per task for all approaches rises since the overall cost for the system also increases with the tasks. However, it is also observed that $CARS-BMW$ and $BMW-PK$ suffer more as we scale the system. The margin between these and optimal approaches grows drastically up to $\sim\$2$. This can be attributed mainly to the increased complexity of the problem as number of tasks is increased and hence the bipartite matching-based heuristic finds less efficient solution compared to optimal. The optimal solutions however have nominal increase in their average price per task ($\sim\$10$) even with scaling compared to rest. 

We also study the effect of scaling V2G tasks to the average final price per task in Fig.~\ref{fig:fin_price_V2G}. We observed similar trend to above but with noticeable gap between optimal and heuristic approaches when only $32\%$ of V2G tasks are available. This results from the sub-optimal performance owing to less number of V2G tasks compared to rest and hence unequal rate of learning the preferences. 

\begin{figure}[htpb]
\minipage{0.49\linewidth}
    \centering
    \includegraphics[width=0.99\linewidth]{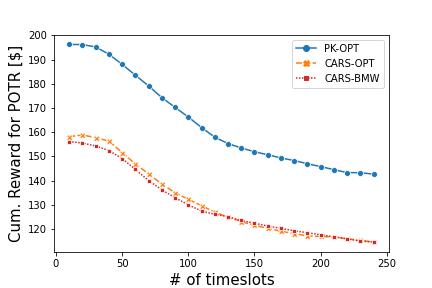}
    \caption{Cumulative reward plot}
    \label{fig:cum_reward}
\endminipage\hfill
\minipage{0.49\linewidth}
    \centering
    \includegraphics[width=0.99\linewidth]{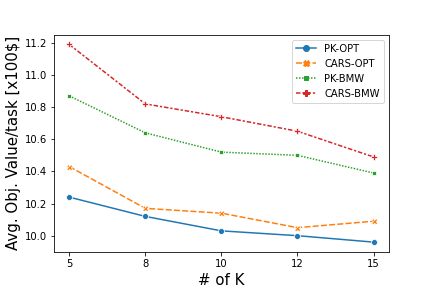}
    \caption{Obj. values/matching vs. K}
    \label{fig:obj_vs_K}
\endminipage
\end{figure}

\noindent 3. \textbf{\underline{Learning accuracy for preferences -- MAE}}: To study the quality of proposed CMAB-based learning algorithm $CARS$ in conjunction with optimal and $BMW$, we use the Mean Absolute Error (MAE) of the learned preferences over time and present them in Fig.~\ref{fig:sMAPE}. Both approaches use same learning algorithm but the solution to $WiBS$ problem differs and thus affects the learning performance. However, this difference is very negligible. Initially, the MAE is $0.28$ and then rapidly decreases to less than $0.05$ for both approaches by $250$ timesteps. The difference in learning efficacy between $CARS-OPT$ and $CARS-BMW$ reduces over the time and is almost same by $250$ timesteps as seen in the graph. Since by $500$ timesteps the system has garnered sufficient knowledge on workers preferences, MAE falls to $0.03$ reflecting the efficacy of proposed CMAB-based preference learning.
Furthemore, we present a cumulative reward plot in Fig.~\ref{fig:cum_reward} that also shows the plots of both learning approaches converge after $200$ timesteps.

\noindent 4. \textbf{\underline{Dependency with $K$}}: In this experiment, we discuss on the dependency of the performance of our proposed approach with recommendation length $K$, as presented in Fig.~\ref{fig:obj_vs_K}. Increasing the number of recommendation $K$ means that the chance of receiving more bids with good quality from same number of workers at the same time increases. This in turn helps to find better solutions which reduce overall cost of the system. This is also verified from the observation in plot of Fig.~\ref{fig:obj_vs_K}. As we increase $K$, the objective values per task over a day's period reduces for all four approaches. Although the perfect and optimal optimal methods do not have significant difference in their performance with varied $K$, the effect is more pronounced in case of bipartite matching based $PK-BMW$ and $CARS-BMW$ where the learning of preferences is benefited by the increased number of bids to choose from with increasing $K$. However, it needs to be noted that pushing $10$ recommends at each timestep can be very intractable for workers and therefore, keeping the length of recommendation list as small as possible is desired.

\section{Conclusion}

e-Uber is a promising crowdsourcing platform for improving the efficiency and sustainability of ride-sharing and energy-sharing services through the use of EVs. It uses reverse auction mechanism to assign spatial tasks to EV drivers based on their preferences, battery level, and other realistic constraints like minimum energy requirement for grid and one-to-one assignment. To optimize the task recommendation process, the platform incorporates user behavioral models including worker preferences and bounded rationality. However, as these preferences are not known \textit{a priori}, e-Uber uses reinforcement learning framework called combinatorial multi-armed bandit for learning the preferences at the runtime based on their feedback. We propose the $CARS$ algorithm that finds optimal solution to both the $POTR$ and $WiBS$ problem. Since the $WiBS$ problem is NP-hard, we propose another bipartite matching-based heuristic, called $BMW$ that finds feasible solution to the winner selection while meeting the minimum V2G energy requirement. Experimental results and simulations demonstrate the effectiveness of e-Uber's approaches, which outperform the baseline algorithm by serving more than 850 tasks within $24$ hours of simulation. On top of that, the baseline often fails to find a feasible solution, rendering it inapplicable in this problem setting. 

Future research could focus on implementing and evaluating e-Uber in real-world settings. This includes the assessment of the impact of different task recommendation and decision prediction algorithms, as well as the integration of new features such as real-time traffic and energy data and dynamic pricing. By exploring these areas, e-Uber has the potential to significantly improve the efficiency and sustainability of ride-sharing and energy-sharing services through the use of EVs.

\section*{Acknowledgment} 
This work is 
supported by the NSF grant EPCN-1936131 and NSF CAREER grant CPS-1943035.

\appendices

\bibliographystyle{IEEEtran}
\bibliography{IEEEabrv,bibliography}

\end{document}